\documentclass[journal, letterpaper]{IEEEtran}

\usepackage{amsmath,amssymb,amsfonts,mathtools}
\usepackage{graphicx}
\usepackage{algorithm}
\usepackage{algorithmic}
\usepackage{booktabs}
\usepackage[numbers]{natbib}
\usepackage[letterpaper,margin=1in]{geometry}
\usepackage{xcolor}
\usepackage{multirow}
\usepackage{xurl}
\usepackage[hidelinks]{hyperref}
\newtheorem{proposition}{Proposition}
\newtheorem{assumption}{Assumption}

\begin{document}

\title{Federated Martingale Posterior Sampling}

\author{
Boning~Zhang,~\IEEEmembership{Student Member,~IEEE,}
Matteo~Zecchin,~\IEEEmembership{Member,~IEEE,}
Mingzhao~Guo,~\IEEEmembership{Student Member,~IEEE,}
Dongzhu~Liu,~\IEEEmembership{Member,~IEEE,}
Osvaldo~Simeone,~\IEEEmembership{Fellow,~IEEE}

\thanks{
Boning Zhang, Mingzhao Guo, and Dongzhu Liu are with the School of Computing, 
University of Glasgow, Glasgow G12 8QQ, U.K. 
(e-mail: b.zhang.6@research.gla.ac.uk; m.guo.2@research.gla.ac.uk; dongzhu.liu@glasgow.ac.uk). Matteo Zecchin is with the Communication Systems Department, EURECOM, 
06904 Sophia Antipolis, France 
(e-mail: Matteo.Zecchin@eurecom.fr). Osvaldo Simeone is with the Institute for Intelligent Networked Systems, 
Northeastern University London, London E1 8PH, U.K. 
(e-mail: o.simeone@northeastern.edu). The work of O. Simeone was supported by the European Research Council (ERC) under the European Union’s Horizon Europe Programme (grant agreement No. 101198347), by an Open Fellowship of the EPSRC (EP/W024101/1), and by the EPSRC project (EP/X011852/1).
}
}

\maketitle

\begin{abstract}
Federated Bayesian neural networks require fixing a prior on the model parameters, which is notoriously difficult, and misspecification of this prior can severely degrade accuracy and calibration. 
Motivated by the rapid progress of predictive models, the martingale posterior, also known as predictive Bayes, replaces the prior--likelihood pair with a predictive distribution and recovers parameter uncertainty by repeatedly drawing predictive samples and refitting the model. 
This letter proposes {federated martingale posterior} (FMP) sampling, a one-shot embarrassingly parallel protocol in which each client uploads a small set of trainable data embeddings and the server runs the predictive sampler centrally. 
Analysis of the sampling error demonstrates the impact of the dataset compression rate, while experiments show that FMP closely matches the centralized counterpart and achieves the lowest mean expected calibration error (ECE) among the evaluated one-shot federated methods.
\end{abstract}

\begin{IEEEkeywords}
Bayesian learning, federated learning, martingale posterior
\end{IEEEkeywords}

\IEEEpeerreviewmaketitle

\section{Introduction}
\label{sec:intro}
Bayesian learning provides a principled framework for quantifying predictive uncertainty of machine learning models~\citep{xu2024bayesian,simeone2022machine, simeone2026uncertainty}, 
while requiring the specification of a prior distribution over a large weight space ~\citep{bissiri2016general,knoblauch2022optimization,zecchin2023robust}.
The rapid progress of foundation models, including large language models, has shown that it is often much easier to obtain a powerful predictor of future observations than to specify a meaningful prior on the parameter space.
This observation has motivated the martingale posterior (MP)~\citep{fong2023martingale}, also known as predictive Bayes~\citep{battiston2025bayesian}, which recovers samples from an implicit posterior over parameters by repeatedly drawing predictive samples and fitting a parametric model via empirical risk minimization (ERM).

\begin{figure}[t]
    \centering
    \includegraphics[width=\linewidth]{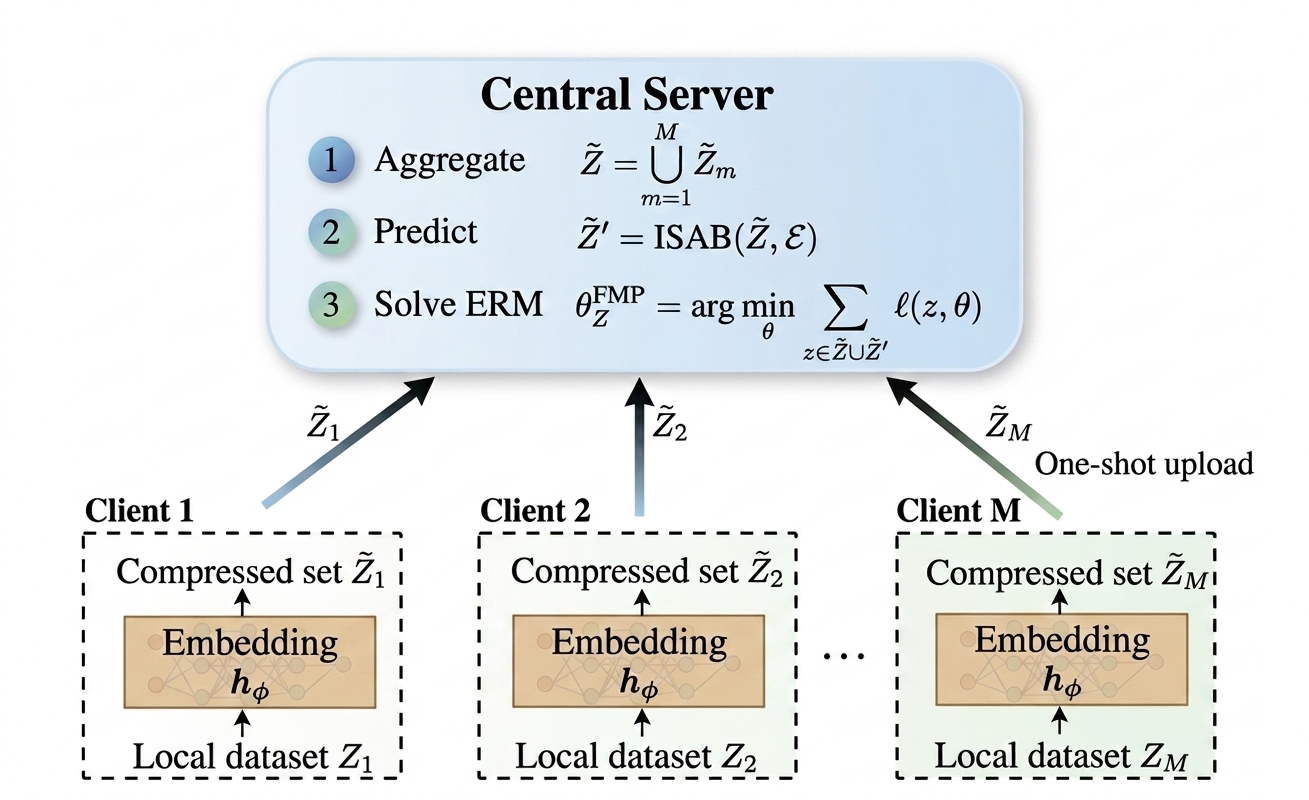}
    \caption{In the proposed FMP protocol, each client compresses its private local dataset \(\mathcal Z_m\) into a compressed set \(\widetilde{\mathcal Z}_m=h_\phi(\mathcal Z_m)\), which is uploaded to the server in one communication round. The server generates predictive samples via induced self-attention blocks (ISAB) and solves the ERM problem to obtain approximate global MP samples.}
    \label{fig:system}
    \vspace{-0.6cm}
\end{figure}

Bayesian learning has also found applications to federated settings~\citep{kassab2022federated, scott2016bayes,
zhu2024federated}.
This letter removes the need to specify parameter-space priors in federated Bayesian learning by proposing federated MP (FMP) sampling, a one-shot embarrassingly parallel protocol that approximates centralized MP from compressed local data. 
The main contributions are as follows:

\noindent $\bullet$ We propose FMP, a one-shot federated sampling protocol based on a shared set-transformer predictor~\citep{lee2019set,lee2023martingale} and on the compression of local datasets into small sets of trainable embeddings (see Fig. \ref{fig:system}).

\noindent $\bullet$ We analyze the sampling error of FMP, relating it to the number of trainable embeddings.

\noindent $\bullet$ We design a meta-training procedure that aligns the FMP samples with those of the centralized MP across a corpus of related tasks, and validate the resulting protocol on several datasets.

\section{Background}
We start by reviewing the martingale posterior (MP) \citep{fong2023martingale}, together with its attention-based implementation described in \cite{lee2023martingale}.
\subsubsection{Martingale Posterior Sampling}
Given a dataset \(\mathcal{Z}=\{z_i\}_{i=1}^n \) with data points \(z_i=(x_i, y_i)\), the MP \citep{fong2023martingale} reframes posterior uncertainty about parameters \(\theta\) as predictive uncertainty on \emph{unseen, hypothetical data} conditional on the observed data. 
Formally, the MP specifies a joint predictive density over unseen data \(\mathcal{Z}' = \{z_i' \}_{i=1 }^{n'}\), denoted by \(p(\mathcal{Z}' | \mathcal{Z})\). 
Given samples \(\mathcal{Z}' \sim p(\mathcal{Z}' | \mathcal{Z})\), MP then solves the empirical risk minimization (ERM) problem  
\begin{equation} \label{equ:mp_general}
\theta^{\text{MP}} = \arg \min_{\theta} \sum_{z \in \mathcal{Z} \cup \mathcal{Z}'} \ell(z, \, \theta) , 
\end{equation}  
where $\ell(z, \, \theta)$ is a loss function. 
Note that the parameters $\theta^{\text{MP}}$ in \eqref{equ:mp_general} are random variables due to the stochasticity of the unseen data $\mathcal{Z}' \sim p(\mathcal{Z}' | \mathcal{Z})$.
The samples $\theta^{\text{MP}}$ are treated as draws from the underlying implicit martingale posterior.


As shown in \cite{fong2023martingale}, assuming that the predictive distribution  \(p(\mathcal{Z}' | \mathcal{Z})\) is conditionally exchangeable (i.e., permutation-invariant) given the dataset \(\mathcal{Z}\),  and choosing the loss \(\ell(z, \theta) = - \log p(z | \theta)\) in \eqref{equ:mp_general}, the distribution of \(\theta^\text{MP}\) converges to the posterior  \(p(\theta | \mathcal{Z})\) as \(n' \rightarrow \infty\) under weak regularity conditions.

\subsubsection{Set Transformer–Based Predictive Distribution}\label{sec:isab}
In \cite{lee2023martingale}, the conditional distribution $p(\mathcal{Z}' | \mathcal{Z})$ is implemented using a set Transformer with induced self-attention blocks (ISAB)~\citep{lee2019set}.
Specifically, the generator first samples an i.i.d. base set $\mathcal{E}=\{\varepsilon_i\}_{i=1}^{n'} \overset{\mathrm{i.i.d.}}{\sim} p(\varepsilon)$, typically from a standard Gaussian distribution $p(\varepsilon) = \mathcal{N}(0,I_d)$.  
Each data point $z_i \in \mathcal{Z}$ is processed by a feedforward neural network $g(\cdot)$ as $g(z_i) \in \mathbb{R}^d$, forming the set $\mathcal{R}=\{g(z_i)\}_{i=1}^n$ of inducing points. 
Then, ISAB produces samples $\mathcal{Z}'$ by applying a cascade of two multi-head attention blocks (MABs) with no masking. 

Denote an MAB block by $\mathrm{MAB}(\mathcal{Q}, \mathcal{C})$, which maps a query set $\mathcal{Q}$ to an updated set by attending to a set $\mathcal{C}$, representing both keys and values, via multi-head attention. 
The first MAB block uses the base set $\mathcal{E}$ and the inducing points $\mathcal{R}$ to produce an intermediate state $H = \operatorname{MAB}(\mathcal{R},\,\mathcal{E})$, and the second MAB block applies attention once more to generate the outputs $\mathcal{Z}' =\operatorname{MAB}(\mathcal{E},\,H)$. 
This architecture can be proved to produce conditionally exchangeable outputs $\mathcal{Z}^\prime = \{z_i^{\prime}\}_{i=1}^{n'}$ \citep{lee2019set}, and its input-output operation is denoted as
\begin{equation} \label{equ:isab}
    \mathcal{Z}' = \mathrm{ISAB}(\mathcal{Z}, \mathcal{E}) = \mathrm{MAB}(\mathcal{E}, \mathrm{MAB}(\mathcal{R}, \mathcal{E})),
\end{equation}
where we recall that the inducing points $\mathcal{R}$ are a function of the data $\mathcal{Z}$.

\section{Problem Formulation and Baseline}
\subsubsection*{A.\quad Setting}
As shown in Fig.~\ref{fig:system}, we consider a setting with $M$ clients, with the $m$-th client holding a private dataset of the size $n$, $\mathcal{Z}_m = \{z_{m,i} = (x_{m, i}, y_{m, i})\}_{i=1}^{n}$. 
The clients are connected to a central server. 
We fix a predictive model $p(\mathcal{Z}'|\mathcal{Z})$, and our goal is to develop distributed protocols that allow the server to draw approximate samples \eqref{equ:mp_general} from the global martingale posterior induced by this predictive module $p(\mathcal{Z}' | \mathcal{Z})$ given the global dataset $\mathcal{Z} = \bigcup_{m=1}^{M} \mathcal{Z}_m$. 

Specifically, we assume a fixed predictive mechanism described by the set transformer \eqref{equ:isab} with a given neural network $g(\cdot)$, which generates inducing points, and a given MAB mechanism $\mathrm{MAB}(\cdot, \cdot)$. The target posterior samples are given by~\eqref{equ:mp_general}, denoted as $\theta^{\text{MP}}_{\mathcal{Z}}$, where $\mathcal{Z}'$ is obtained via \eqref{equ:isab}.
We are specifically interested in embarrassingly parallel schemes, in which the clients communicate with the server only once.

\subsubsection*{B.\quad Consensus Federated Martingale Posterior Sampling}\label{sec:cfmp}
As a simple baseline approach, one can directly apply the consensus Monte Carlo protocol introduced in \cite{scott2016bayes}, obtaining a benchmark that we refer to as the Consensus Federated Martingale Posterior (CFMP). In the CFMP setup, each client $m$ generates local unseen data $\mathcal{Z}_m^\prime$ using the predictive distribution $p(\mathcal{Z}_m'| \mathcal{Z}_m)$ based on the local data $\mathcal{Z}_m$. Then, the client solves problem~\eqref{equ:mp_general}, obtaining the samples $\theta_m^{\mathrm{MP}}$ from the local martingale posterior.

The samples $\theta^{\mathrm{MP}}_{m}$ are transmitted to the server, which aggregates all the local samples $\{\theta_m^{\mathrm{MP}}\}_{m=1}^{M}$ to approximate a sample $\theta_{\mathcal{Z}}^{\mathrm{MP}}$ from the global martingale posterior based on the full dataset $\mathcal{Z}$. 
Specifically, CFMP applies the weighted sum 
\begin{equation}\label{equ:cfmp}
    \hat{\theta}^{\mathrm{CFMP}}_{\mathcal{Z}}
    = \Bigg(\sum_{m=1}^{M} \hat{\Sigma}_m^{-1}\Bigg)^{-1} \sum_{m=1}^{M} \hat{\Sigma}_m^{-1} \theta_m^{\mathrm{MP}}, 
\end{equation}
where the covariance matrix $\hat{\Sigma}_m$ is an estimate of the true covariance matrix $\Sigma_m = \mathbb{E}\!\big[(\theta_m^{\mathrm{MP}}-\mathbb{E}[\theta_m^{\mathrm{MP}}])(\theta_m^{\mathrm{MP}}-\mathbb{E}[\theta_m^{\mathrm{MP}}])^\top\big]$  obtained using past samples $\theta_m^{\mathrm{MP}}$.

\section{Federated Martingale Posterior Sampling}\label{sec:fmps}
Constructing the global martingale posterior requires each client $m$ to transmit its local dataset $\mathcal{Z}_m$ to the server.
To avoid this raw-data exposure and reduce bandwidth requirements, this section introduces FMP, a novel federated learning protocol that replaces all uses of the local data $\mathcal{Z}_m$ at the server with trainable embeddings.

\subsubsection{The FMP Protocol}
As illustrated in Fig.~\ref{fig:system}, the FMP protocol leverages an embedding function $h_\phi(\cdot)$ shared among clients, which takes as input a local dataset $\mathcal{Z}_m$ to produce a compressed dataset $\widetilde{\mathcal{Z}}_m=\{\tilde{z}_{m, i}\}_{i=1}^s$ with $s < n$ fictitious data points $\tilde{z}_{m, i} \in \mathbb{R}^d$ for $i=1, \cdots, s$. 
FMP reduces only the cardinality of the local set, from $n$ to $s$, while preserving the dimension $d$ of each observation.
This way, the hyperparameter $s$ controls the communication cost, with a smaller value of $s$ implying a lower communication cost. 
The design of the embedding function is discussed in Sec.~\ref{sec:embed_func}.

The server aggregates the compressed datasets as a surrogate for the original dataset $\mathcal{Z}$ as 
\(
\widetilde{\mathcal{Z}} = \bigcup_{m=1}^{M} \widetilde{\mathcal{Z}}_m.
\)
Using this compressed dataset in lieu of the original dataset, the server draws predictive samples $\widetilde{\mathcal{Z}}'$ using the predictive model \eqref{equ:isab}, i.e.,
\(
\widetilde{\mathcal{Z}}'=\operatorname{ISAB}(\widetilde{\mathcal{Z}}, \mathcal{E}),
\)
where $\mathcal{E}=\{\varepsilon_i\}_{i=1}^{n'}$ is an i.i.d. base set as defined in Sec.~\ref{sec:isab}.
Then, the server obtains approximate martingale posterior samples $\theta_{\mathcal{Z}}^{\mathrm{FMP}}$ by solving the problem
\begin{equation} \label{equ:fmp}
\theta^{\text{FMP}}_{\mathcal{Z}} = \arg \min_{\theta} \sum_{z \in \widetilde{\mathcal{Z}} \cup \widetilde{\mathcal{Z}}'} \ell(z, \, \theta) . 
\end{equation} 

In general, due to the discrepancy between the local datasets $\{\mathcal{Z}_m\}_{m=1}^{M}$ and the compressed datasets $\{\widetilde{\mathcal{Z}}_m\}_{m=1}^M$, the distribution of the samples $\theta^{\text{FMP}}_{\mathcal{Z}}$ differs from that of the martingale posterior samples $\theta^{\text{MP}}_{\mathcal{Z}}$. To mitigate this issue, we propose a methodology to optimize the parameters $\phi$ of the embedding function $h_{\phi}(\cdot)$ in Sec.~\ref{sec:meta-training}.

\subsubsection{Embedding Function}\label{sec:embed_func}
The embedding function is instantiated using the Pooling by Multihead Attention (PMA) block \citep{lee2019set}. 
Building on the MAB block, PMA takes a \emph{learnable} set of $s$ seed vectors $\mathcal{S}_{\phi} \in \mathbb{R}^{s \times d}$ as queries, while keys and values are obtained from the local dataset $\mathcal{Z}_m$. 
This is done by applying a per-data point feedforward layer $f_{\phi}(\cdot)$ to each element of the dataset $\mathcal{Z}_m$. 
Overall, the compressed set $\widetilde{\mathcal{Z}}_m = \{\tilde{z}_{m, i}\}_{i=1}^s$ is given by
\begin{equation}\label{equ:embedding}
    \widetilde{\mathcal{Z}}_m = h_\phi(\mathcal{Z}_m) = \mathrm{MAB}(\mathcal{S}_{\phi}, f_{\phi}(\mathcal{Z}_m)).
\end{equation}

\subsubsection{Meta-training the Embedding Function}\label{sec:meta-training}
In order to optimize the embedding function $h_\phi(\cdot)$, we assume the server has access to a meta-training corpus of tasks $\{\mathcal{T}^i\}$ drawn i.i.d. from a distribution \(p(\mathcal{T})\). 
By exposing the embedding function $h_\phi(\cdot)$ to different label subsets and client partitions during meta-training, FMP is designed to generalize to unseen target tasks, following standard meta-learning practice \citep{finn2017model,triantafillou2020meta}.

Each task $\mathcal T^i$ is associated with local datasets $\{\mathcal Z_m^i\}_{m=1}^{M}$, whose pooled dataset is $\mathcal Z^i=\bigcup_{m=1}^{M}\mathcal Z_m^i$.
The data in our federated setting represents a new task $\mathcal{T}^{\text{new}} = \{\mathcal{Z}_m\}_{m=1}^{M}$ sampled from the same distribution  $p(\mathcal{T})$.

Given a per-task base set $\mathcal{E}^i$ used in the predictive mechanisms \eqref{equ:isab} to produce the unseen data $\mathcal{Z}'$ and $\widetilde{\mathcal{Z}}'$, we define the \emph{per-task loss} as
\begin{equation}
\ell^i(\phi;\mathcal{E}^i) \;=\; \|\theta_{\mathcal{Z}^{i}}^{\mathrm{MP}} -\,\theta_{\mathcal{Z}^{i}}^{\mathrm{FMP}} \|,
\label{eq:per_task_loss}
\end{equation}
where $\theta_{\mathcal{Z}^{i}}^{\mathrm{MP}}$ and $\theta_{\mathcal{Z}^{i}}^{\mathrm{FMP}}$ are the samples obtained by the centralized MP scheme and by FMP, respectively. 
The objective \eqref{eq:per_task_loss} compares samples $\theta_{\mathcal{Z}^{i}}^{\mathrm{MP}}$ and $\theta_{\mathcal{Z}^{i}}^{\mathrm{FMP}}$ using the same dataset $\mathcal{Z}^{i}$ and base set $\mathcal{E}^{i}$, and it can be readily seen to yield an unbiased estimate of an upper bound on the 1-Wasserstein distance $W_1(\cdot,\cdot)$ between the conditional distributions of the samples produced by MP and FMP~\citep{gabriel2019computational}.

For meta-training, we sample $K$ tasks \(\{\mathcal{T}^{i}\}_{i=1}^{K}\) i.i.d. from the distribution $p(\mathcal{T})$, along with their corresponding i.i.d. base sets $\{\mathcal{E}^i\}_{i=1}^{K} \overset{\mathrm{i.i.d}}{\sim} p(\varepsilon)$. 
The meta-training objective is defined as the empirical average
\begin{equation}\label{equ:meta-method}
\mathcal{L}(\phi) = \frac{1}{K} \sum_{i=1}^K  \ell^i(\phi;\mathcal{E}^i)=
\frac{1}{K} \sum_{i=1}^K 
\|\theta_{\mathcal{Z}^{i}}^{\mathrm{MP}} -\,\theta_{\mathcal{Z}^{i}}^{\mathrm{FMP}} \|,
\end{equation}
where we recall that the parameters $\phi$ determine both the set of seed vectors $\mathcal{S}_\phi$ and the per-sample transformation $f_\phi(\cdot)$. 
During meta-training, only the parameters $\phi$ of the embedding function $h_{\phi}$ are updated via gradient descent.

\subsubsection{Sampling Approximation Guarantee}\label{sec:sampling_error}

For a nonempty finite dataset $\mathcal{A}=\{a_i\}_{i=1}^{N}$, let $P_{\mathcal{A}}=\sum_{i=1}^N \delta_{a_i}/N$ denote the corresponding empirical probability measure, where $\delta_{a_i}$ is the Dirac delta at $a_i$.
The following assumption posits that similar input datasets yield similar unseen data through the ISAB generator in~\eqref{equ:isab}, with similarity measured by the 1-Wasserstein distance $W_1(\cdot,\cdot)$~\citep{gabriel2019computational}.

\begin{assumption}[Generator stability]
\label{ass:generator_stability}
Given the base set $\mathcal{E}$, there exists a constant $C\ge0$ such that the predictive generator in~\eqref{equ:isab} satisfies the inequality
\begin{equation}
W_1\!\left(
P_{\mathrm{ISAB}(\mathcal{Z}_1,\mathcal{E})},
P_{\mathrm{ISAB}(\mathcal{Z}_2,\mathcal{E})}
\right)
\le C \cdot W_1(P_{\mathcal{Z}_1},P_{\mathcal{Z}_2})
\label{eq:generator_stability}
\end{equation}
for any two input datasets $\mathcal{Z}_1$ and $\mathcal{Z}_2$.
\end{assumption}

\begin{proposition}[FMP approximation bound]
\label{prop:fmp_approximation}
Assume that the loss $\ell(z,\theta)$ is $\mu$-strongly convex in $\theta$ and $L_z$-Lipschitz in $z$, uniformly with respect to the other argument, and that Assumption~\ref{ass:generator_stability} holds.
Then, for a shared base set $\mathcal{E}$ and for any compression factor $s/n$, with $1 \le s \le n$, we have the bound
\begin{equation}
\begin{aligned}
&\big\|\theta_{\mathcal{Z}}^{\mathrm{MP}}
      -\theta_{\mathcal{Z}}^{\mathrm{FMP}}\big\|\\
&\quad \le
2\left\{
\frac{L_z}{\mu}
\left[
\bigl(b_s+(1-b_s)C\bigr)\delta_s
+\Delta_s D
\right]
\right\}^{1/2},
\end{aligned}
\label{eq:fmp_approximation_bound}
\end{equation}
where $b_s=Ms/(Ms+n')$; $\Delta_s=Mn/(Mn+n')-b_s$; $\delta_s=W_1(P_{\mathcal{Z}},P_{\widetilde{\mathcal{Z}}})$ measures the discrepancy between the pooled original dataset $\mathcal{Z}$ and the pooled compressed dataset $\widetilde{\mathcal{Z}}$; 
and $D=W_1(P_{\mathcal{Z}},P_{\mathcal{Z}'})$ measures the discrepancy between the pooled original dataset and its predictive samples $\mathcal{Z}'=\mathrm{ISAB}(\mathcal{Z},\mathcal{E})$.

The bound~\eqref{eq:fmp_approximation_bound} expresses the error between the centralized MP sample $\theta_{\mathcal{Z}}^{\mathrm{MP}}$ and the FMP sample $\theta_{\mathcal{Z}}^{\mathrm{FMP}}$ as a function of the compression ratio $s/n$ via two terms.
The first term accounts for the error caused by replacing the original data with compressed data in the training loss, while the second captures the predictive error. 
When $s=n$ and $\widetilde{\mathcal{Z}}=\mathcal{Z}$, both terms vanish, and FMP recovers centralized MP.
\end{proposition}

\section{Experiments and Conclusions}
\subsubsection{Baselines}
In this section, we consider three classes of protocols, namely \emph{local}, \emph{centralized}, and \emph{one-shot federated (embarrassingly parallel)} protocols. 
Within each class, methods are further categorized as frequentist or Bayesian. 
For Bayesian methods, we adopt standard ensembling using model parameter samples.

{Local} protocols operate independently at each client using only local data, including the artificial neural network (LANN), Bayesian neural network (LBNN), and martingale posterior (LMP) baselines. 
Conversely, centralized protocols operate on the pooled dataset formed by all clients, yielding the ANN, BNN, and MP baselines \citep{fong2023martingale}.
As one-shot federated benchmarks, we include consensus artificial neural network (CANN) as the frequentist baseline, in which the server constructs a global model through a single averaging step over the locally trained parameters \citep{NIPS2010_abea47ba}; consensus Bayesian neural network (CBNN), in which the server combines local posterior samples using consensus Monte Carlo \citep{scott2016bayes}; FedLPA, which communicates layer-wise Laplace posterior approximations \citep{NEURIPS2024_9482f45f}; Bayesian pseudocoresets (BPC), which communicate optimized pseudo-observations in function space \citep{dhondt2024pseudocoresets}; CFMP (see Sec.~\ref{sec:cfmp} ); and the proposed FMP (Sec.~\ref{sec:fmps}).

\subsubsection{Classification Tasks}
We evaluate all methods on MNIST, CIFAR-10, and a 20-class subset of CIFAR-100, allowing us to assess performance across varying levels of task complexity. 
We first consider a setting in which all clients share the same label space and data are evenly distributed across clients. 
We then study a heterogeneous setting in which data are partitioned using a Dirichlet distribution with concentration parameter \(\alpha \in \{0.1, 0.5, 1.0, 5.0\}\)~\citep{hsu2019measuring}.
Rather than operating on raw images, we first map each input image \(x\) to a frozen feature representation \(h \in \mathbb{R}^{d_x}\), obtained from a feature extractor trained offline with cross-entropy loss.

\subsubsection{Experimental Results}

\begin{table}[t]
\centering
\caption{Accuracy (ACC, \%) and ECE under homogeneous client partitions on MNIST, CIFAR-10, and a 20-class CIFAR-100 subset, with per-client communication costs reported for CIFAR-10 only.
Results are reported as mean $\pm$ standard deviation over three seeds for ACC and ECE.}
\vspace{-5pt}
\label{tab:iid_main}
\setlength{\tabcolsep}{2pt}
\renewcommand{\arraystretch}{1.1}
\resizebox{\linewidth}{!}{%
\begin{tabular}{@{}c cc@{\hspace{6pt}}cc@{\hspace{6pt}}cc c@{}}
\toprule
Method
& \multicolumn{2}{c}{MNIST}
& \multicolumn{2}{c}{CIFAR-10}
& \multicolumn{2}{c}{CIFAR-100}
& \shortstack{Comm.\\(KB)} \\
\cmidrule(lr){2-3}
\cmidrule(lr){4-5}
\cmidrule(lr){6-7}
& ACC $\uparrow$ & ECE $\downarrow$
& ACC $\uparrow$ & ECE $\downarrow$
& ACC $\uparrow$ & ECE $\downarrow$
& \\
\midrule
LANN
& $95.17\pm0.18$ & $0.071\pm0.011$
& $57.43\pm0.15$ & $0.085\pm0.013$
& $57.36\pm0.16$ & $0.180\pm0.031$
& -- \\
LBNN
& $95.10\pm0.21$ & $0.079\pm0.021$
& $55.95\pm0.13$ & $0.082\pm0.017$
& $56.26\pm0.17$ & $0.184\pm0.024$
& -- \\
LMP
& $96.43\pm0.31$ & $0.068\pm0.012$
& $68.96\pm0.27$ & $0.105\pm0.021$
& $66.25\pm0.29$ & $0.199\pm0.033$
& -- \\
\midrule
ANN
& $97.24\pm0.22$ & $0.052\pm0.010$
& $78.63\pm0.25$ & $0.073\pm0.015$
& $70.30\pm0.26$ & $0.050\pm0.009$
& -- \\
BNN
& $97.13\pm0.42$ & $0.045\pm0.017$
& $78.36\pm0.36$ & $0.059\pm0.022$
& $70.10\pm0.35$ & $0.039\pm0.008$
& -- \\
MP
& $97.33\pm0.23$ & $0.021\pm0.004$
& $80.57\pm0.15$ & $0.034\pm0.003$
& $70.95\pm0.18$ & $0.027\pm0.005$
& -- \\
\midrule
CANN
& $96.89\pm0.28$ & $0.068\pm0.013$
& $74.06\pm0.23$ & $0.213\pm0.044$
& $68.40\pm0.26$ & $0.316\pm0.053$
& 5.12 \\
CBNN
& $97.05\pm0.29$ & $0.062\pm0.019$
& $73.57\pm0.33$ & $0.161\pm0.025$
& $69.50\pm0.36$ & $0.246\pm0.078$
& 51.20 \\
FedLPA
& $96.93\pm0.44$ & $0.064\pm0.010$
& $73.48\pm0.36$ & $0.171\pm0.033$
& $69.22\pm0.24$ & $0.263\pm0.043$
& 38.36 \\
BPC
& $96.95\pm0.28$ & $0.051\pm0.020$
& $76.32\pm0.34$ & $0.107\pm0.031$
& $69.81\pm0.33$ & $0.136\pm0.022$
& 15.48 \\
CFMP
& $97.00\pm0.23$ & $0.053\pm0.009$
& $77.40\pm0.19$ & $0.142\pm0.025$
& $69.73\pm0.28$ & $0.222\pm0.019$
& 51.20 \\
FMP
& $97.05\pm0.19$ & $0.041\pm0.012$
& $78.91\pm0.21$ & $0.057\pm0.025$
& $70.32\pm0.22$ & $0.042\pm0.023$
& 15.48 \\
\bottomrule
\end{tabular}%
}
\vspace{-6pt}
\end{table}

Table~\ref{tab:iid_main} reports the classification accuracy (ACC) and expected calibration error (ECE) \citep{guo2017calibration} under homogeneous client partitions on MNIST, CIFAR-10, and the 20-class CIFAR-100 subset.
Among all the one-shot federated methods, FMP achieves the lowest mean ECE on all three datasets, while attaining the highest or tied-highest mean accuracy.
Moreover, FMP remains close to the centralized MP benchmark despite using compressed data-level embeddings.
Table~\ref{tab:iid_main} also reports the per-client communication cost for all one-shot federated methods under the CIFAR-10 setting.

\begin{figure}[t]
    \centering
    \includegraphics[width=0.95\linewidth]{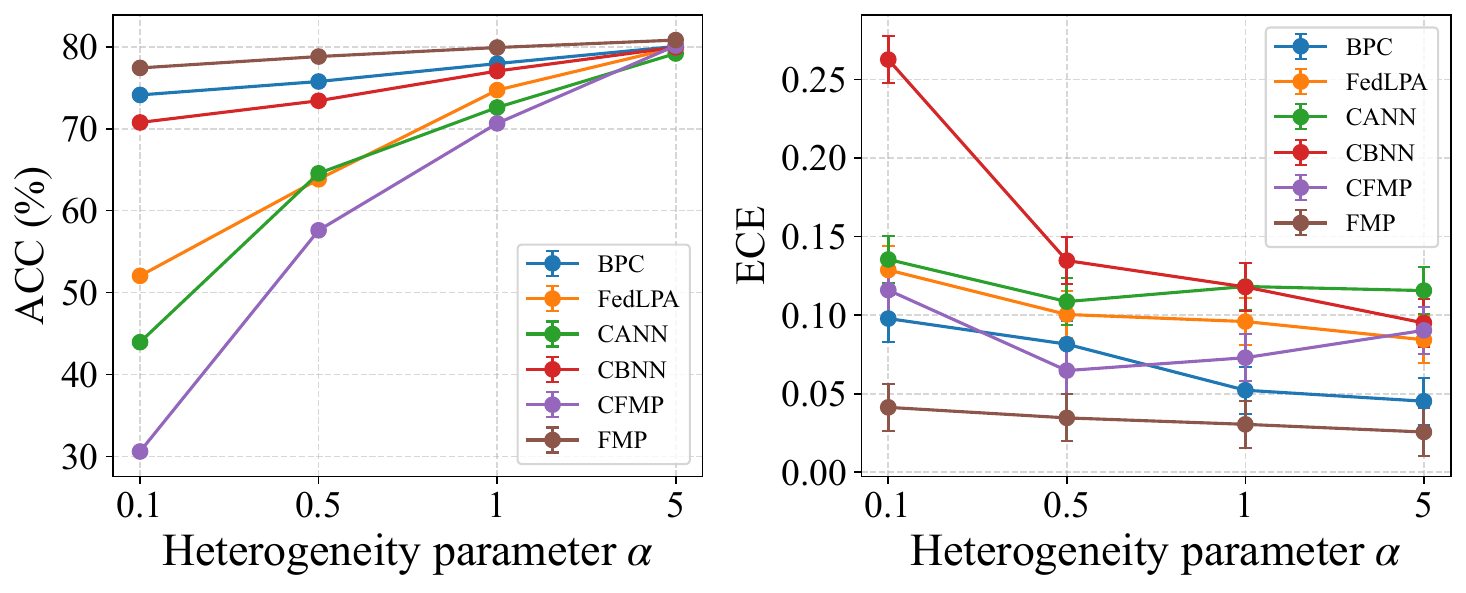}
    \vspace{-10pt}
    \caption{Accuracy (ACC) and ECE on CIFAR-10 under heterogeneous client partitions with Dirichlet parameters $\alpha \in \{0.1,0.5,1.0,5.0\}$. Left: ACC versus Dirichlet $\alpha$. Right: ECE versus Dirichlet $\alpha$. Error bars denote standard deviations over three seeds.}
    \vspace{-0.5cm}
    \label{fig:niid_results}
\end{figure}

Fig.~\ref{fig:niid_results} reports the accuracy and ECE under heterogeneous client partitions.
FMP achieves the highest mean accuracy and the lowest mean ECE across all considered values of $\alpha$, with the largest advantage observed under the strongest heterogeneity at $\alpha=0.1$.

\begin{figure}[t]
    \centering
    \includegraphics[width=\linewidth]{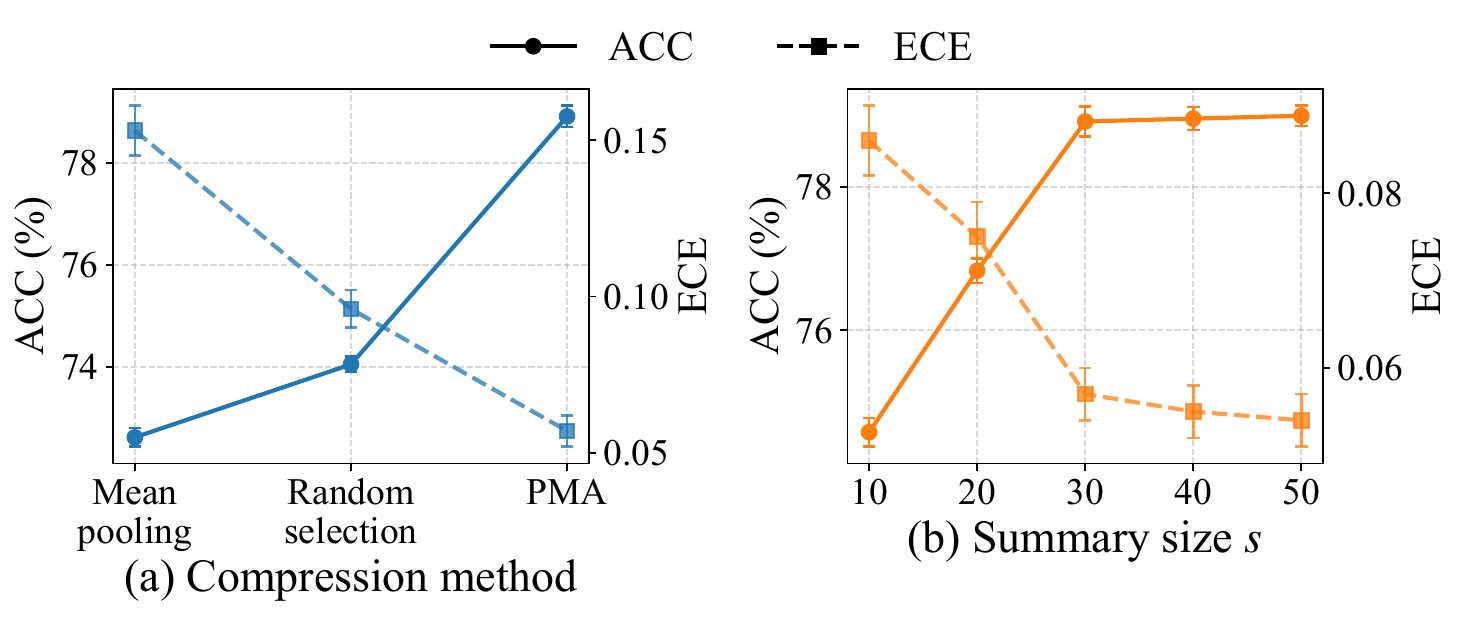}
    \vspace{-20pt}
    \caption{Compression ablation on CIFAR-10 under homogeneous client partitions. (a) Comparison of compression methods with a fixed summary size $s=30$. (b) Effect of the PMA summary size $s$. Error bars denote standard deviations over three seeds.}
    \vspace{-0.5cm}    
    \label{fig:ablation}
\end{figure}

\subsubsection{On The Impact of Compression}
We evaluate the effects of the compression operator and summary size on CIFAR-10 under homogeneous client partitions.
At a fixed summary size $s=30$, we compare PMA with random selection and mean pooling under the same communication budget. 
All three compression schemes use the same predictive sampler and classifier, so that only the compression operator is varied.
We further vary $s$ for PMA to study the communication--performance trade-off.
As shown in Fig.~\ref{fig:ablation}, PMA achieves higher accuracy and lower ECE than the two non-learned compression schemes, while the performance gains become marginal beyond $s=30$.

\subsubsection{Conclusions}
Overall, FMP closely tracks the centralized MP benchmark and improves calibration over the evaluated one-shot federated baselines, especially under heterogeneous client partitions.
Future work may study formal privacy guarantees, adaptive communication-accuracy tradeoffs, and variants that are robust to communication errors.

\bibliographystyle{IEEEtran}
\bibliography{refs}

@article{xu2024bayesian,
  title={Bayesian deep learning via expectation maximization and turbo deep approximate message passing},
  author={Xu, Wei and Liu, An and Zhang, Yiting and Lau, Vincent},
  journal={IEEE Trans. Signal Process.},
  volume={72},
  pages={3865--3878},
  year={2024},
}

@book{simeone2022machine,
  title={Machine learning for engineers},
  author={Simeone, Osvaldo},
  year={2022},
  publisher={Cambridge University Press},
}

@article{simeone2026uncertainty,
  title={Uncertainty quantification and data efficiency in {AI}: {An} information-theoretic perspective},
  author={Simeone, Osvaldo and Romano, Yaniv},
  journal={IEEE Journal on Selected Areas in Information Theory},
  year={2026},
  publisher={IEEE}
}

@article{bissiri2016general,
  title={A general framework for updating belief distributions},
  author={Bissiri, Pier Giovanni and Holmes, Chris C and Walker, Stephen G},
  journal={J. Roy. Statist. Soc. Ser. B},
  volume={78},
  number={5},
  pages={1103--1130},
  year={2016},
}

@article{knoblauch2022optimization,
  title={An optimization-centric view on {Bayes}' rule: {Reviewing} and generalizing variational inference},
  author={Knoblauch, Jeremias and Jewson, Jack and Damoulas, Theodoros},
  journal={J. Mach. Learn. Res.},
  volume={23},
  number={132},
  pages={1--109},
  year={2022},
}

@article{zecchin2023robust,
  title={Robust {PAC}$^{m}$: {Training} ensemble models under misspecification and outliers},
  author={Zecchin, Matteo and Park, Sangwoo and Simeone, Osvaldo and Kountouris, Marios and Gesbert, David},
  journal={IEEE Trans. Neural Netw. Learn. Syst.},
  volume={35},
  number={11},
  pages={16518--16532},
  year={2023},
}

@article{fong2023martingale,
  title={Martingale posterior distributions},
  author={Fong, Edwin and Holmes, Chris and Walker, Stephen G},
  journal={J. Roy. Statist. Soc. Ser. B},
  volume={85},
  number={5},
  pages={1357--1391},
  year={2023},
}

@article{battiston2025bayesian,
  author  = {Battiston, Marco and Cappello, Lorenzo},
  title   = {{Bayesian} predictive inference beyond martingales},
  journal = {arXiv:2507.21874},
  year    = {2025},
}

@article{kassab2022federated,
  title={Federated generalized {Bayesian} learning via distributed {Stein} variational gradient descent},
  author={Kassab, Rahif and Simeone, Osvaldo},
  journal={IEEE Trans. Signal Process.},
  volume={70},
  pages={2180--2192},
  year={2022},
}

@article{scott2016bayes,
  title={{Bayes} and big data: {The} consensus {Monte Carlo} algorithm},
  author={Scott, Steven L and Blocker, Alexander W and Bonassi, Fernando V and Chipman, Hugh A and George, Edward I and McCulloch, Robert E},
  journal={International Journal of Management Science and Engineering Management},
  volume={11},
  number={2},
  pages={78--88},
  year={2016},
  publisher={Taylor \& Francis}
}

@article{zhu2024federated,
  title={Federated inference with reliable uncertainty quantification over wireless channels via conformal prediction},
  author={Zhu, Meiyi and Zecchin, Matteo and Park, Sangwoo and Guo, Caili and Feng, Chunyan and Simeone, Osvaldo},
  journal={IEEE Trans. Signal Process.},
  volume={72},
  pages={1235--1250},
  year={2024},
}

@inproceedings{lee2019set,
  title={{Set Transformer}: {A} framework for attention-based permutation-invariant neural networks},
  author={Lee, Juho and Lee, Yoonho and Kim, Jungtaek and Kosiorek, Adam and Choi, Seungjin and Teh, Yee Whye},
  booktitle={Proc. Int. Conf. Mach. Learn. (ICML)},
  pages={3744--3753},
  year={2019},
}

@inproceedings{lee2023martingale,
  title={Martingale posterior neural processes},
  author={Lee, Hyungi and Yun, Eunggu and Nam, Giung and Fong, Edwin and Lee, Juho},
  booktitle={Proc. Int. Conf. Learn. Represent. (ICLR)},
  year={2023}
}

@inproceedings{finn2017model,
  title={Model-Agnostic Meta-Learning for fast adaptation of deep networks},
  author={Finn, Chelsea and Abbeel, Pieter and Levine, Sergey},
  booktitle={Proc. Int. Conf. Mach. Learn. (ICML)},
  pages={1126--1135},
  year={2017},
}

@inproceedings{triantafillou2020meta,
  title={{Meta-Dataset}: {A} dataset of datasets for learning to learn from few examples},
  author={Triantafillou, Eleni and Zhu, Tyler and Dumoulin, Vincent and Lamblin, Pascal and Evci, Utku and Xu, Kelvin and Goroshin, Ross and Gelada, Carles and Swersky, Kevin and Manzagol, Pierre-Antoine and Larochelle, Hugo},
  booktitle={Proc. Int. Conf. Learn. Represent. (ICLR)},
  year={2020},
}

@article{gabriel2019computational,
  title={Computational optimal transport with applications to data sciences},
  author={G. Peyr{\'e} and M. Cuturi},
  journal={Foundations and Trends{\textregistered} in Machine Learning},
  volume={11},
  number={5-6},
  pages={355--607},
  year={2019},
  publisher={Emerald Publishing Limited}
}

@inproceedings{NIPS2010_abea47ba,
 author = {Zinkevich, Martin and Weimer, Markus and Li, Lihong and Smola, Alex},
 booktitle = {Proc. Adv. Neural Inf. Process. Syst.},
 pages = {2595--2603},
 title = {Parallelized Stochastic Gradient Descent},
 volume = {23},
 year = {2010}
}

@inproceedings{NEURIPS2024_9482f45f,
 author = {Liu, Xiang and Liu, Liangxi and Ye, Feiyang and Shen, Yunheng and Li, Xia and Jiang, Linshan and Li, Jialin},
 booktitle = {Proc. Adv. Neural Inf. Process. Syst.},
 pages = {81510--81548},
 title = {{FedLPA}: {One-shot} federated learning with layer-wise posterior aggregation},
 volume = {37},
 year = {2024}
}

@article{dhondt2024pseudocoresets,
  title={One-shot federated learning with {Bayesian} pseudocoresets},
  author={D'Hondt, Tim and Pechenizkiy, Mykola and Peharz, Robert},
  journal={arXiv:2406.02177},
  year={2024},
}

@article{hsu2019measuring,
  title={Measuring the effects of non-identical data distribution for federated visual classification},
  author={Hsu, Tzu-Ming Harry and Qi, Hang and Brown, Matthew},
  journal={arXiv:1909.06335},
  year={2019},
}

@inproceedings{guo2017calibration,
  title={On calibration of modern neural networks},
  author={Guo, Chuan and Pleiss, Geoff and Sun, Yu and Weinberger, Kilian Q},
  booktitle={Proc. Int. Conf. Mach. Learn. (ICML)},
  pages={1321--1330},
  year={2017},
}

\end{document}